\documentclass[10pt,twocolumn,letterpaper]{article}

\usepackage{iccv}
\usepackage{times}
\usepackage{epsfig}
\usepackage{graphicx}
\usepackage{amsmath}
\usepackage{amssymb}
\usepackage[numbers]{natbib}
\usepackage{multirow}
\usepackage{tablefootnote}
\usepackage{adjustbox}
\usepackage[table,xcdraw]{xcolor}

\usepackage[breaklinks=true,bookmarks=false]{hyperref}

\iccvfinalcopy 

\begin{document}

\title{ Encoding Binary Concepts in the Latent Space of Generative Models for Enhancing Data Representation}

\author{Zizhao Hu\\
University of Southern California\\
{\tt\small zizhaoh@usc.edu}
\and
Mohammad Rostami\\
University of Southern California\\
{\tt\small rostamim@usc.edu}
}

\maketitle

\begin{abstract}
    Binary concepts\footnote{``Concepts" are also known as ``factors'' in unsupervised learning literature. We use these terms interchangeably.} are empirically used by humans to generalize efficiently. And they are based on Bernoulli distribution which is the building block of information. These concepts span both low-level and high-level features such as  ``large vs small" and ``a neuron is active or inactive''. Binary concepts are ubiquitous features and can be used to transfer knowledge to improve model generalization. We propose a novel binarized regularization to facilitate learning of binary concepts to improve the quality of data generation in autoencoders. We introduce a binarizing hyperparameter $r$ in data generation process to disentangle the latent space symmetrically. We demonstrate that this method can be applied easily to existing variational autoencoder (VAE) variants to encourage symmetric disentanglement, improve reconstruction quality, and prevent posterior collapse without computation overhead. We also demonstrate that this method can boost existing models to learn more transferable representations and generate more representative samples for the input distribution which can alleviate catastrophic forgetting using generative replay under continual learning settings.

\end{abstract}

\section{Introduction}

There is a famous game named Akinator. It asks you a series of binary answer questions such as ``is he young or old?'' or ``is she an actress'' to find out what subject you are thinking of based on your answers. A recent study \cite{akinator} suggests humans use a similar strategy to ask questions with binary answers to maximize information gain. This implies that humans use binary concepts to efficiently encode information internally to discriminate between concepts. In reinforcement learning, there is evidence \cite{GalilAI} that an agent can generalize low-level physical properties by sequentially binarizing observations. The above evidence serves as an inspiration to benefit from binarized concepts to learn latent space representations for an input distribution such that they disentangle concepts in a binary scheme to potentially facilitate   learning   all factors. We especially apply our approach on VAEs since they learn   stable latent representations.

\begin{figure} 
    \centering
    \includegraphics[width=0.47\textwidth]{ 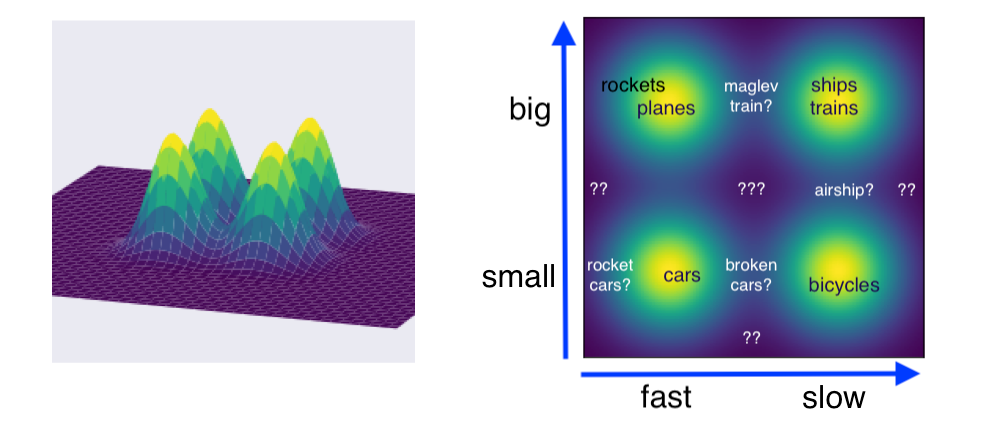}
    \caption{Two binary concepts represented by a mixture of two symmetric Gaussians in each dimension of a latent space with dimension $d=2$. It has $2^d$ global density maxima.}
    \label{fig:mesh1}
\end{figure} 

Most existing variants of VAEs focus on regularizing the latent space to learn more disentangled factors to boost efficiency and the quality of generated data. Considering that (i) low-level information-dense factors are difficult to disentangle, and are often treated as noise (posterior collapse) and (ii) high-level information-scarce factors are easy to disentangle, they usually lead to learning low-variance representations. As a result, generated samples are highly biased toward the mean of the learned distribution in the latent space and hence are less diverse compared to the real-world data.
To our knowledge,  existing methods are not exploring the possibility of learning an inherent property shared by all levels of factors in real-world data.  We argue that binary concepts in the latent representation are a suitable candidate for this purpose (see Figure~\ref{fig:mesh1}). 

We introduce a general regularization technique to improve VAEs. We  call it   ``binarized" regularization  because it enforces learning latent distributions that approximate the posterior using two symmetric modalities in each dimension of the latent space (see Figure \ref{fig:mesh1}). As a result, the latent distribution mass deterministically moves towards one of the closest prior distributions on a high-dimensional grid of a mixture distribution that allows generating diverse samples (note Figure \ref{fig:mesh1} is an example for a 2D  latent embedding space).
We demonstrate that our regularization technique improves disentanglement of the latent and simultaneously improves reconstruction quality through extensive experiments.
Our specific contributions include:

\begin{itemize}
    
\item  We develop a   regularization technique to binarize the VAE objective and demonstrate that binarized VAEs are able to further disentangle the latent, resolve noisy low-level factors in generated images, and can generate significantly more variant high-level features.

\item  We demonstrated that our method can prevent posterior collapsing by restoring the collapsed latent representation to learn low-level factors. 

\item  We further adopt our idea to address catastrophic forgetting in continual learning. Results demonstrate that our method is able to generate more diverse samples for the past tasks and compares favorably against state-of-the-art generative replay methods on the most challenging incremental-class classification tasks.
\end{itemize}

\section{Background and Related Work}

\begin{figure}
    \centering
    \includegraphics[width =8cm]{ 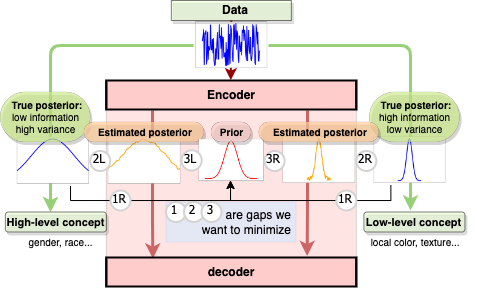}
    \caption{Latent optimization process. We only show two out of infinite true posteriors here to simplify the representation.}
    \label{fig:vae_pipe}
\end{figure}

\paragraph{Variational Autoencoders} have an extremely rich literature. We propose a unified framework to formulate all current VAE-based methods in Figure~\ref{fig:vae_pipe}. We argue that a Gaussian Mixture Model (GMM) can be used to model any complicatedly-distributed latent concepts using a combination of an infinite number of simply-distributed high-level   to low-level concepts. These concepts continuously span from high-variance to low-variance. The gaps in Figure~\ref{fig:vae_pipe} represent how close the distributions are between
\textbf{Gap 1:} prior and true posterior\footnote{the optimal mapping from data to a regularized distribution}, \textbf{Gap 2 (L/R):} approximate posterior and true posterior, and \textbf{Gap 3 (L/R):} approximate posterior and prior. A smaller Gap 2 leads to a better reconstruction quality. A smaller Gap 3 leads to better disentanglement. The goal of the VAE optimizations can be generalized as reducing Gap 3 (preferably without reducing Gap 2). Within this formulation, the prominent VAE variants can be categorized by the way they achieve this goal:

\textbf{(i) Reducing the size of Gap 1:} 
These methods indirectly reduce both Gaps 2 and 3 on certain types of true posteriors.
\textbf{Gaussian VAE} \cite{vae} uses a unimodal Gaussian distribution as the prior. It can serve as a good approximation for many true posteriors but the generated samples are biased towards its mean and may not be very diverse. 
\textbf{Hyperspherical VAE (NVAE)}  \cite{svae} uses von Mises-Fisher (vMF)   prior that is closer to some specific true posteriors.

\textbf{(ii) Reducing the size of Gap 3 by trading-off the size of Gap 2:}
\textbf{$\beta$-VAE} \cite{bvae} uses a hyperparameter $\beta$ to do so.

\textbf{(iii) Reducing the size of Gap 3 with adaptable priors:}
\textbf{GMVAE} \cite{gmvae} creates a finite set of candidate priors.
\textbf{disentangling $\beta$-VAE}\cite{dbvae} creates an infinite set of candidate priors on an ``equal KL-divergence'' line (see Figure \ref{fig:priors} row 2 column 1 all dashed curves for clarity). 

\textbf{(iv) Reducing the size of Gap 3 by adapting the prior to approximate the posterior:}
\textbf{VAMP} \cite{vamp} has a mixture prior, and uses aggregate posterior\footnote{The observed latent distribution on aggregate seen data.} to move prior towards the approximated posterior during training.
\textbf{$\beta$-TCVAE} \cite{btcvae} adds a regularization term to optimize the prior towards the posterior during training.
\textbf{DIP-VAE} \cite{dipvae} uses the concept of ``moment'' to predict the future posterior, and uses this extra information to move the prior towards the approximate posterior. \textbf{FactorVAE}  \cite{factorvae} adds a regularization term to factorize prior to adapt to low-variance posterior.

\textbf{(v) Reducing the size of Gap 2R and Gap 3R with a low-variance local prior:}
To achieve this goal without increasing Gap 2L and 3L, having many low-variance approximate posteriors is necessary to approximate the high-variance distribution. These posteriors result in an inaccurate approximation of hte high-level factors when a model is small. The extreme case is \textbf{Autoencoders} \cite{ae}.
\textbf{WAE} \cite{wae} only maintains prior-like global approximate posteriors, but reduces the variance of local posteriors.
\textbf{VQVAE} \cite{vqvae} nears the extreme with a no-variance quantized latent prior. 

\textbf{(vi) Reducing the size of Gap 2 and Gap 3 with larger models:}
A larger model can reduce Gap 2 and Gap 3 by providing an approximate posterior with a larger variance range.   \textbf{VDVAE}\cite{vdvae} is an example in this category.

\textbf{(vii) Reducing the size of Gap 2 with a less regularized likelihood prior:}
This approach is not within the VAE literature   since it does not reduce Gap 3, but it is worth mentioning it due to its compatibility with this framework. The reconstruction prior used in VAEs is local low-variance Bournourli distribution. This requirement can be relaxed to more global distributions seen in \textbf{GAN} \cite{gan}, or distributed throughout the full model as in \textbf{diffusion models} \cite{ddpm}. 

Our   method employs   ideas of (i) and (iii). When used on a unimodal Gaussian VAE, it is a GMVAE with $2^d$ components, with deterministic  assignment conditioned on the approximate posterior, and unimodal Gaussian reparameterization to distributed the mass in the latent embedding.

 \paragraph{Continual Learning} aims to learn a set of sequentially arriving tasks using a shared model. When the model is updated to learn the current task,  we observe catastrophic forgetting~\cite{kirkpatrick2017overcoming} in the model due to deviation from the optimal parameters for the past learned tasks. 
CL algorithms employ three primary strategies to mitigate catastrophic forgetting. One approach involves regularizing a fixed shared model, such that different information pathways are used to learn each task's weights~\cite{kirkpatrick2017overcoming,aljundi2018memory,jin2021learn}. The aim is to identify critical model parameters that encode the learned knowledge of each task and consolidate these parameters while updating the model to learn new tasks. The downside is that the learning capacity of the model is compromised as more weights are consolidated. Another approach relies on model expansion~\cite{rusu2016progressive,yoonlifelong}, which involves adding new weights to a base model and customizing the network to learn new tasks via these additional weights. The downside is that the model can grow unboundedly as more tasks are learned. Finally, most algorithms employ pseudo-rehearsal through experience replay~\cite{rolnick2019experience,rostami2021detection,rostami2021lifelong}. This method involves storing a representative subset of training data for each task in a memory buffer and replaying those samples along with the current task's data to maintain encoded knowledge about previously learned. A group of algorithms relaxes the need for a memory buffer by building a generative model that learns to generate pseudo-samples that are very similar to the real samples of the past learned tasks. We demonstrate that our approach helps to improve this category of CL methods.

\section{Problem Description}
\subsection{Variational Autoencoders Objective}

Given a dataset $\mathcal{D} = \{x_i\}_{i=1}^n$,  VAEs learn an encoder $p_{\phi}(z|x)$ and a decoder $p_{\theta}(x|z)$ that maximize the likelihood $p_{\phi,\theta}(x)$ to approximate the empirical evidence $p_{\mathcal{D}}(x)$. The encoder   maps   $p_{\mathcal{D}}(x)$ to a simple latent prior $p(z)$ which has better sampling properties. The objective is to maximize:
\begin{align*}
\ln p_{\phi,\theta}(x)  \geq \mathbb{E}_{p_{\phi}(z|x)}[\ln p_{\theta}(x|z)] - D_{KL} [p_{\phi}(z|x) || p(z)],
\end{align*}
where the right hand side is ELBO, served as the proxy objective we maximize to learn the latent distribution or   as the objective in the following minimization problem:
\begin{equation}
\mathcal{L} = \underbrace{-\mathbb{E}_{p_{\phi}(z|x)}[\ln p_{\theta}(x|z)]}_\text{Reconstruction Loss} + \underbrace{D_{KL} [p_{\phi}(z|x) || p(z)]}_\text{KL Loss},
\label{eq1}
\end{equation}
where the first term is the negative likelihood of data after passing through the model. It is   modeled as the reconstruction loss in VAEs. The second term is the KL-divergence between the approximate posterior and the latent prior.

\subsection{Shortcoming of VAEs}
The two terms in Eq.~\eqref{eq1} conflict with each other. The first term trades disentanglement for reconstruction quality, emphasizing it results in low interpretability and control of latent representation. The second term trades reconstruction quality for disentanglement, emphasizing that it results in   low variances of high-level concepts that lead to a high correlation with latent dimensions and high noise levels of low-level concepts that lead to posterior collapse. We develop a method to to mitigate these challenges.

\section{Proposed Solution}

\subsection{Binarized Modification}

We define a new binarized objective that instead of minimizing the distributional distance between the approximate posterior and the prior minimizes the closest distributional distance between the approximate posterior and two binarized priors which are located $r$ away from the mean of the prior. The new binarized loss   can be written as:

\begin{equation}
\begin{aligned}  
\mathcal{L}_\mathcal{B} = &-\mathbb{E}_{p_{\phi}(z|x)}[\ln p_{\theta}(x|z)] \\&+\mathbf{1}_{\mu_{p_{\phi}(z|x)}>\mu_{p(z)}} D [p_{\phi}(z|x) || p(z-r)]\\
&+ \mathbf{1}_{\mu_{p_{\phi}(z|x)}<\mu_{p(z)}} D [p_{\phi}(z|x) || p(z+r)],
\end{aligned}
\end{equation}
where $r \in [0, +\infty]$ is a model hyperparameter. It decides how disentangled the two candidate distributions are (see Figure~\ref{fig:mesh1}). Note that the loss reduces to the original VAE loss when $r = 0$. A more general form can be written as: 
\begin{equation}
\begin{aligned}  
\mathcal{L}_\mathcal{B} = \mathcal{L}(|\mu_{p(z)}-\mu_{p_{\phi}(z|x)}|-r),
\end{aligned}
\end{equation}
for any loss function $\mathcal{L}(\mu_{p(z)}-\mu_{p_{\phi}(z|x)})$, where $\mu_{p(z)}$ is a location variable of a prior distribution, and $\mu_{p_{\phi}(z|x)}$ is a location variable of an approximate posterior distribution.

\subsection{Binarized Gaussian Prior}
\textbf{Gaussian Prior Closed-form Loss}

To add the binarized modification to a unimodal Gaussian prior $p(z)\sim \mathcal{N}(\mu,\sigma^2)$, we have the closed form: 
\begin{equation}
\begin{aligned}   
\mathcal{L}_\mathcal{B} = &-\mathbb{E}_{p_{\phi}(z|x)}[\ln p_{\theta}(x|z)] - \frac{1}{2}\\
&+\ln(\frac{\sigma_{p_{\phi}(z|x)}}{\sigma_{p(z)}}) + \frac{\sigma_{{p_{\phi}(z|x)}}^2 + (|\mu_{p_{\phi}(z|x)} - \mu_{p(z)}| - r)^2}{2\sigma_{p(z)}^2} 
\end{aligned}
\label{eq2}
\end{equation}

 We can show that this is equivalent to ELBO with a binarized $z$-dependent prior as the following:
 
\begin{equation} \label{eq22}
p_B (z)= 
\begin{cases}
   \mathcal{N}(\mu_{p(z)}+r,\, \sigma_{p(z)}^2),
   & \text{if } \mu_{p_{\phi}(z|x)} - \mu_{p(z)}\geq 0\\\\
   \mathcal{N}(\mu_{p(z)}-r,\, \sigma_{p(z)}^2), &\text{otherwise}
\end{cases}
\end{equation}

\begin{figure}
    \includegraphics[width=0.98\linewidth]{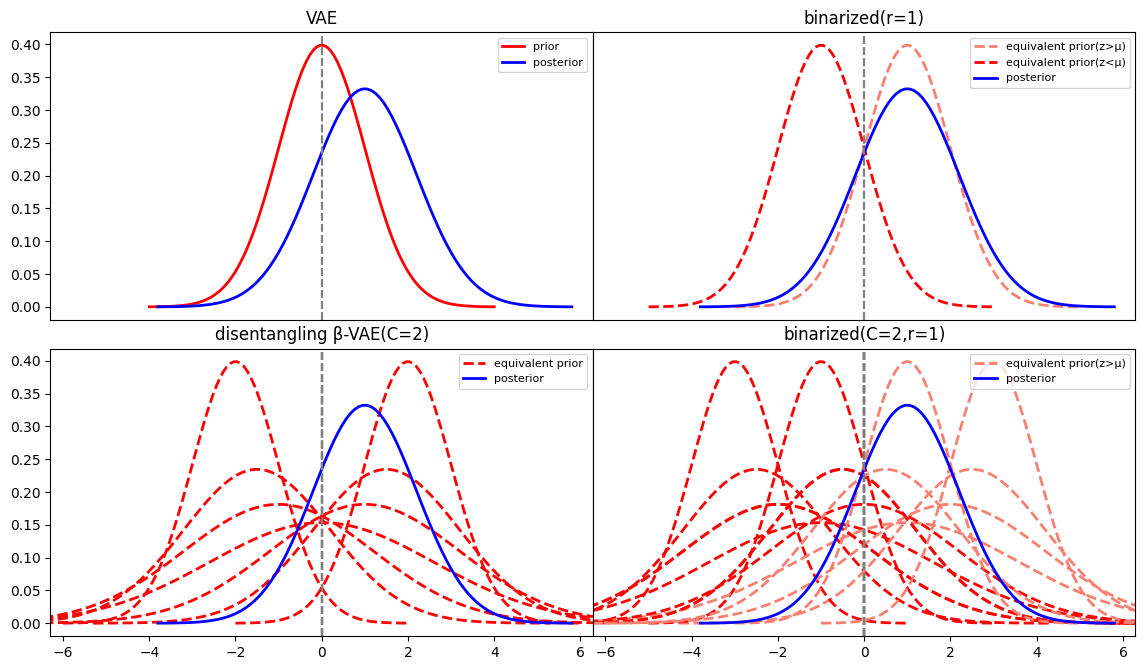}
    \caption{Equivalent priors for binarized models.}
    \label{fig:priors}
\end{figure}

This equivalent prior can be visualized as in Figure~\ref{fig:priors}. For more details on derivation please refer to the Appendix.

\subsection{Reparametrization}
We still use the original reparametrization trick since we still consider the original prior once the estimated posterior is decided. This assumption maintains the continuous property of the approximate posterior at the prior mean which is broken by the deterministic assignment. The probabilistic reparametrization and the deterministic assignment create an instability close to the prior mean, which can alleviate posterior collapse. The source of the instability is the opposing force provided by the Bernoulli-distribution-like encoding. It will have   maximum information capacity when posterior collapse happens which discourages collapse.

\subsection{Sampling}

The instability at the prior mean will force the latent to deviate from the high probability density center. This allows more samples from the underrepresented outer region of high-variance factors, and fewer samples from the biased center high-density region, achieving reduced-bias sampling. Although this will negatively impact low-variance factors, in practice, the combined effect achieves better total sampling quality. This might be due to the alleviated posterior collapse compensating for this negative impact.

\begin{figure}[h]
    \centering
    \includegraphics[width=0.5\textwidth]{ 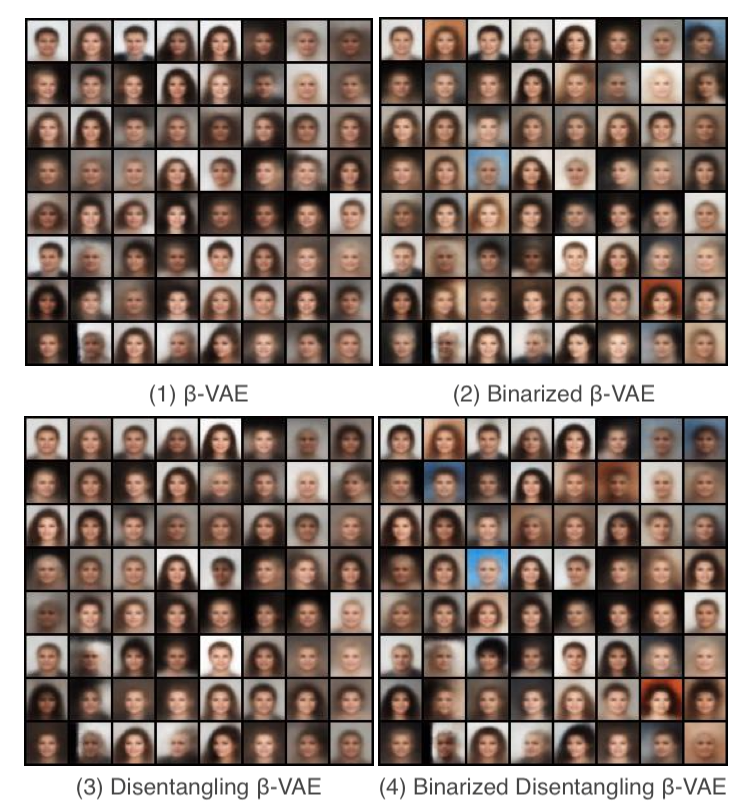}
    \caption{Reconstruction visualization}
    \label{fig:recon4model}
\end{figure}
\begin{figure}[h]
    \centering
    \includegraphics[width=7cm, height=4.375cm]{ 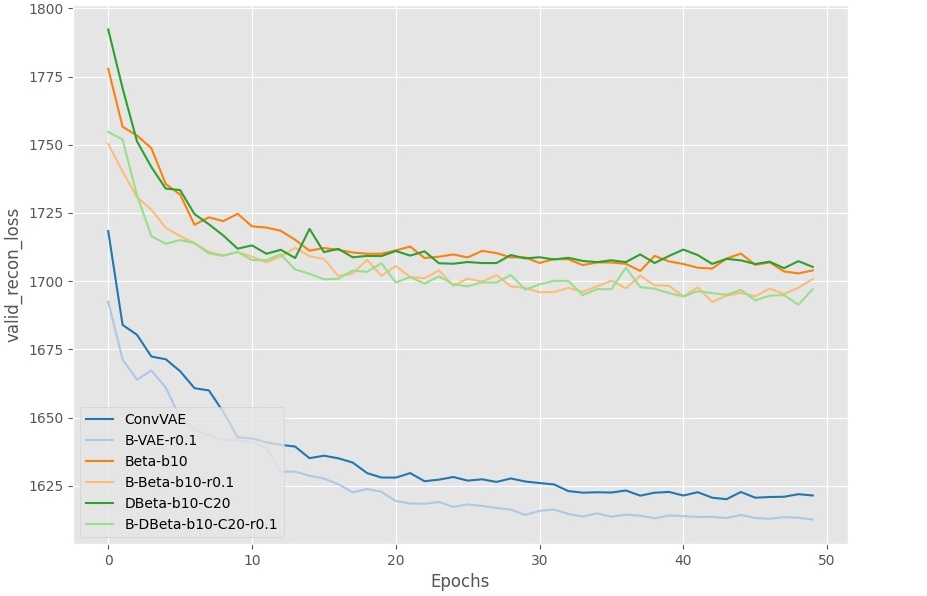}
    \caption{Validation reconstruction loss. 
  With ``B-" meaning binarized, ``(D)BETA" meaning (disentangling)$\beta$-VAE, ``r",``b",``C" are binarizing, $\beta$-VAE, and disentangling $\beta$-VAE hyperparameters $r$, $\beta(\gamma)$, and $C$, followed by the values.}
    \label{fig:curve4model}
\end{figure}

\textbf{Reduced-bias Sampling:}
For   the two binarized distributions, we assign an equal probability to sample from:
\[P(z_{\text{sample}}) = \frac{1}{2}\mathcal{N}(\mu_{p(z)}+r,\, \sigma_{p(z)}^2)\\ + \frac{1}{2} \mathcal{N}(\mu_{p(z)}-r,\, \sigma_{p(z)}^2)
\]
which is approximately sampling from the   Gaussian:
\[\mathcal{N}(\mu_{p(z)},\, \sigma_{p(z)}^2)\]
when $r$ is small.  In all of our experiments, this sampling method is used due to its simplicity and nice properties.

\textbf{Biased Sampling:}
If we want to sample from the   distribution when the input dataset is imbalanced,
we restore the distribution of the dataset by conditioning the sampling prior on the latent statistics of the full dataset:
\begin{equation*}
\begin{aligned}
p(z> \mu_{p(z)}|X) \mathcal{N}(\mu_{p(z)}+r,\, \sigma_{p(z)}^2) \\
+ p(z<\mu_{p(z)}|X) \mathcal{N}(\mu_{p(z)}-r,\, \sigma_{p(z)}^2)
\end{aligned}
\end{equation*}
This sampling would for a skewed distribution.

\section{Empirical Results}
\textbf{Experimental set-up:}
we used four types of  VAE  models: 
\textbf{toyVAE} is a feed-forward VAE, used to test theories and conduct ablation tests on the MNIST dataset. 
\textbf{ConvVAE} is a medium-size convolutional VAE, used for analytical experiments on MNIST and CelebA(32x32, 64x64) datasets.
\textbf{L-ConvVAE} is a large-size convolutional VAE, used for qualitative visualization experiments on CelebA( 64x64) dataset. 
\textbf{Cl-VAE} is used for continual learning on splitMNIST, permutedMnist, and CIFAR100 datasets. All comparisons   have identical model architecture, training settings, and data preprocessing. For   details and implementation,   refer to supplementary material.

\subsection{Reconstruction and Unsupervised Clustering}

First, we use toyVAE as the base model to compare the performance of binarized VAE with VAE on the MNIST dataset. Reconstruction is evaluated using image-sum pixel-wise binary cross-entropy (BCE). Unsupervised clustering is evaluated by normalized mutual information (NMI). The result is presented in Table \ref{table:MNIST}. We observe that binarized VAE is able to achieve better reconstruction and clustering performances with different latent emebdding sizes. 

\begin{table*}[h]
\centering
\resizebox{0.9\textwidth}{!}{%
\begin{tabular*}{1.01\linewidth}{|c|c|cccc|cccc|ccc|}
\hline
                    &    & \multicolumn{4}{c|}{BCE}                                                    & \multicolumn{4}{c|}{NMI}                                                          & \multicolumn{3}{c|}{training time(s)}  \\ \hline
epochs              & d  & \multicolumn{1}{c|}{AE}      & VAE    & NVAE             & BVAE             & \multicolumn{1}{c|}{AE}     & VAE             & NVAE            & BVAE            & VAE          & NVAE       & BVAE       \\ \hline
\multirow{5}{*}{10} & 2  & \multicolumn{1}{c|}{136.19}  & 138.72 & 136.83           & \textbf{135.66*} & \multicolumn{1}{c|}{0.546*} & 0.505           & 0.535           & \textbf{0.536}  & $\sim$10     & $\sim$13   & $\sim$10   \\
                    & 4  & \multicolumn{1}{c|}{111.84*} & 114.01 & 114.33           & \textbf{112.83}  & \multicolumn{1}{c|}{0.675}  & 0.657           & 0.647           & \textbf{0.695*} & $\sim$10     & $\sim$17   & $\sim$10   \\
                    & 8  & \multicolumn{1}{c|}{87.59}   & 91.91  & 92.29            & \textbf{86.32*}  & \multicolumn{1}{c|}{0.700}  & 0.701           & 0.696           & \textbf{0.713*} & $\sim$11     & $\sim$20   & $\sim$11   \\
                    & 16 & \multicolumn{1}{c|}{69.23}   & 85.46  & 81.52            & \textbf{67.85*}  & \multicolumn{1}{c|}{0.691}  & 0.700           & 0.671           & \textbf{0.746*} & $\sim$14     & $\sim$24   & $\sim$14   \\
                    & 32 & \multicolumn{1}{c|}{57.26}   & 84.26  & 97.37            & \textbf{56.15*}  & \multicolumn{1}{c|}{0.624}  & \textbf{0.715*} & 0.608           & 0.668           & $\sim$18     & $\sim$29   & $\sim$18   \\ \hline
\multirow{5}{*}{20} & 2  & \multicolumn{1}{c|}{134.03}  & 135.43 & \textbf{132.91*} & 134.32           & \multicolumn{1}{c|}{0.530}  & 0.565           & \textbf{0.578*} & 0.556           & \multicolumn{3}{c|}{\multirow{5}{*}{}} \\
                    & 4  & \multicolumn{1}{c|}{109.36*} & 112.05 & 110.61           & \textbf{109.81}  & \multicolumn{1}{c|}{0.691}  & 0.670           & 0.677           & \textbf{0.712*} & \multicolumn{3}{c|}{}                  \\
                    & 8  & \multicolumn{1}{c|}{84.32}   & 88.85  & 88.65            & \textbf{84.00*}  & \multicolumn{1}{c|}{0.731}  & 0.738           & 0.719           & \textbf{0.746*} & \multicolumn{3}{c|}{}                  \\
                    & 16 & \multicolumn{1}{c|}{65.63}   & 83.27  & 76.70            & \textbf{64.87*}  & \multicolumn{1}{c|}{0.749}  & 0.695           & 0.740           &\textbf{0.754*}          & \multicolumn{3}{c|}{}                  \\
                    & 32 & \multicolumn{1}{c|}{50.15*}  & 81.25  & 86.2             & \textbf{50.40}   & \multicolumn{1}{c|}{0.650}  & \textbf{0.700*} & 0.631           & 0.668           & \multicolumn{3}{c|}{}                  \\ \hline
\end{tabular*}%
}
\caption{Reconstruction and clustering result. Numbers with an Asterisk denote the best performance overall. Numbers in bold denote the best performance excluding auto-encoder. (Additional columns: Autoencoder(AE), hyperspherical VAE(NVAE), training time; Additional rows: training epoch, latent dimensionality($d$) are provided for ablation test in the last section. $d<8$ converges at epoch 10.)}
\label{table:MNIST}
\end{table*}

Next, we scale the model to ConvVAE. We test VAE, $\beta$-VAE, disentangling $\beta$-VAE, and their binarized modifications (6 models) on the CelebA dataset (32x32 face-aligned). We again observe that the binarized models showed lower reconstruction loss (Figures \ref{fig:recon4model} and \ref{fig:curve4model}), lower variational loss, and higher ELBO compared to the base models.

\subsection{Analyzing Reconstruction and Sampling}
\begin{figure}[ht]
\includegraphics[width=0.48\textwidth]{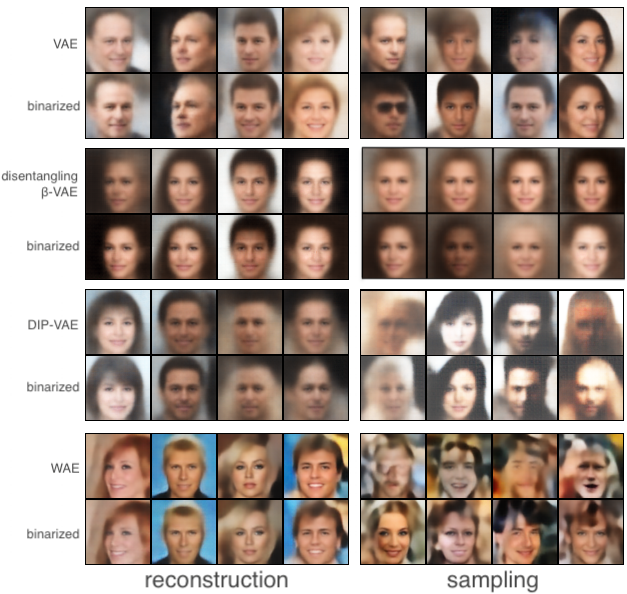}
    \caption{Reconstruction and sampling visualization. Binarized models lead to more variant and realistic samples.}
    \label{fig:recon_samples}
\end{figure}

We then further scale up the base model to L-ConvVAE. We test VAE, $\beta$-VAE, disentangling $\beta$-VAE, DIP-VAE, WAE and their binarized modifications on the CelebA (64x64 face-aligned) dataset, and visualize the reconstruction and sampling results in Figure \ref{fig:recon_samples}. We use the same sampling latent space between the original and binarized models.

\begin{figure}[h]
    \centering
\includegraphics[width=0.4\textwidth]{ 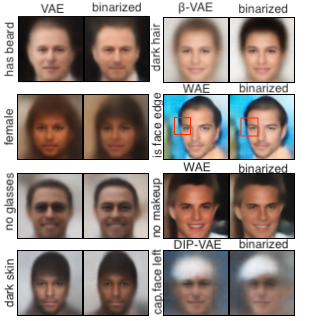}
    \caption{Reconstruction of samples with underrepresented or ambiguous properties}
\label{fig:binary_concepts}
\end{figure}

For reconstruction, the binarized versions have sharper edges and more detailed facial features in general and are more accurate in reconstructing those underrepresented categories in CelebA dataset such as old, facial hair, wearing a cap, etc. (Figure \ref{fig:binary_concepts}).
For sampling, we observe that low-level properties are similar between the samples, while high-level properties are well-distinctive and more variant in their binarized versions. This observation aligns well with our hypothesis. An interesting observation is that when binarizing WAE, the binarized model is able to generate samples with significantly more realistic facial features. 

\subsection{Resolving Posterior Collapsing}
\begin{figure}[h]
    \centering
\includegraphics[width=0.45\textwidth]{ 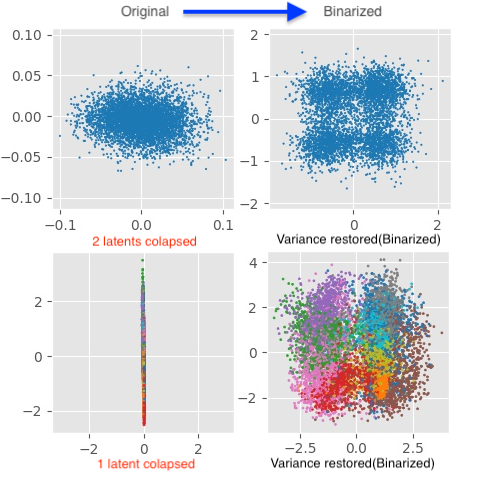}
    \caption{ \small Binarized modification reused  collapsed latent dimensions. Shown are 2 out of 16 latent spaces in disentangling $\beta$-VAE. Row 1 is CelebA(32x32 face-aligned) and Row 2 is  MNIST.}
    \label{fig:restore_var}
\end{figure}

\begin{figure}[h]
    \centering
    \includegraphics[width=0.5\textwidth]{ 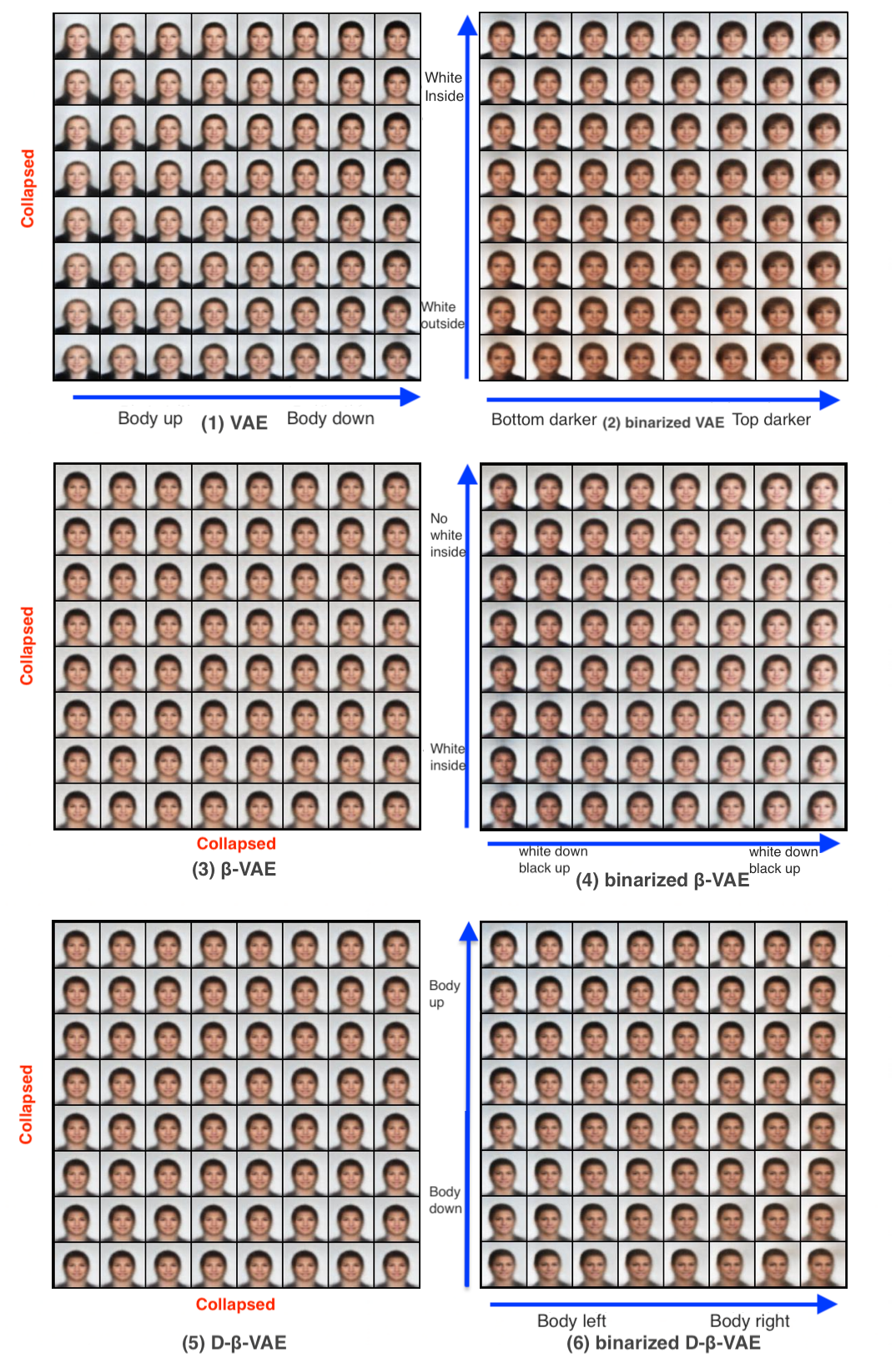}
    \caption{Bottom-2 variance latent interpolation using the CelebA dataset. The binarized models learn low-level properties, while the original models collapse}
    \label{fig:lowvar}
\end{figure}

We use ConvVAE as the base model to visualize the latent space for MNIST and CelebA(32x32, 64x64 face-aligned) reduced-size(60000 instances) datasets. We tested VAE, $\beta$-VAE, disentangling $\beta$-VAE, and the binarized models. All the base models suffer from posterior collapse due to the adjusted smaller dataset compared to the large capacity of the ConvVAE decoder. We see that the binarized models are able to maintain the collapsed latent dimensions (see Figures \ref{fig:restore_var} and \ref{fig:lowvar}), by introducing instability at the prior mean.
concept  (see Figure \ref{fig:restore_var}).

\begin{figure}[h]
    \centering
    \includegraphics[width=0.5\textwidth]{ 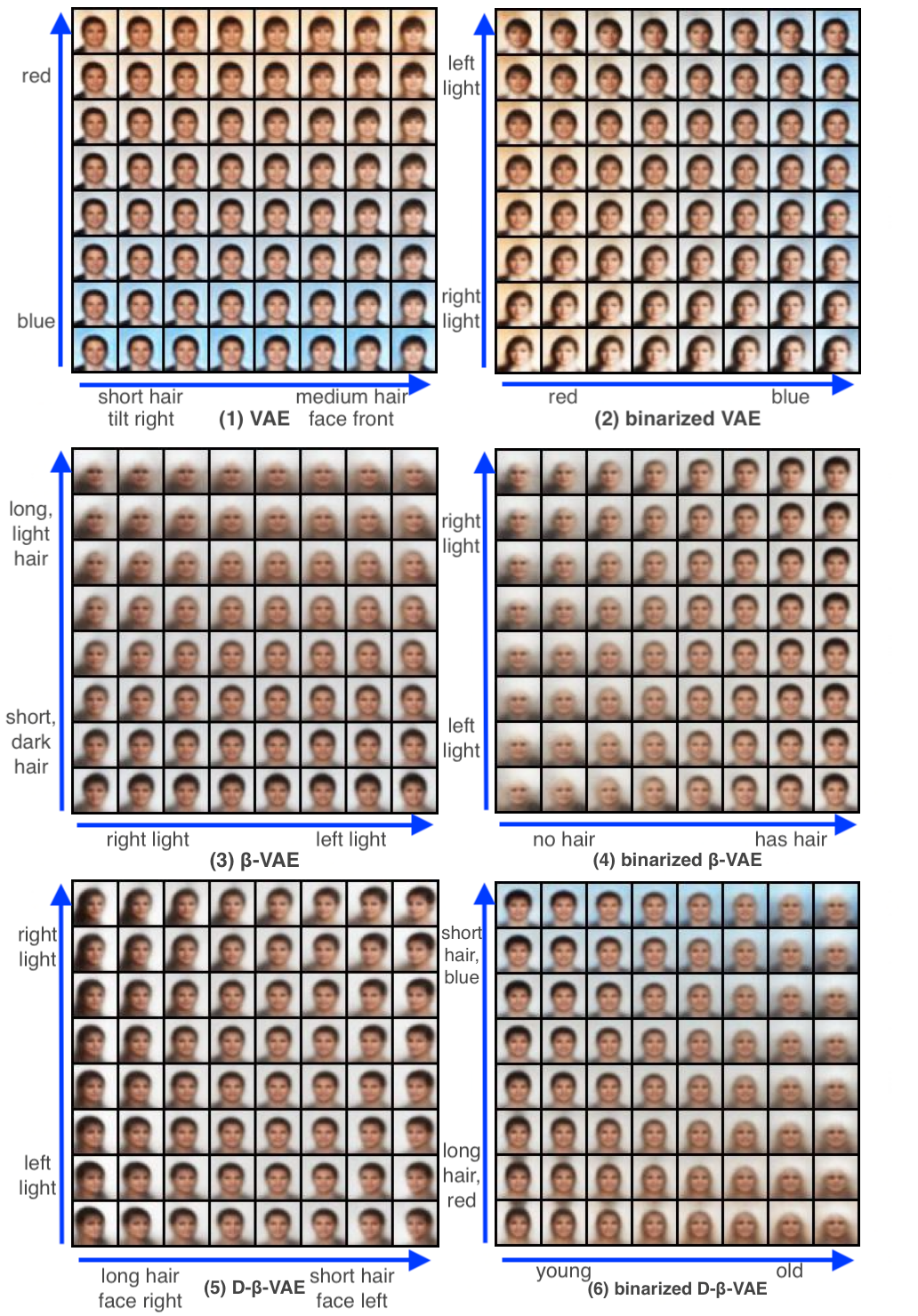}
    \caption{Top-2 variance latent interpolation. The binarized models learn high variance, efficient binary concepts}
    \label{fig:highvar}
\end{figure}

Then we look at what exact concepts are learned in the high-variance and low-variance latent dimensions. We see more variations of low-level binary concepts are learned by the binarized models in the high-variance dimensions  (see Figure \ref{fig:highvar}).
While high-level concepts are learned in the low-variance dimensions restored by the binarized models. While these concepts are unlearnable by the original VAE models (see Figure \ref{fig:lowvar}).

\subsection{Generative Replay in Continual Learning}

\begin{figure}[h]
    \centering
\begin{minipage}{.23\textwidth}
  \centering
  \includegraphics[width=\linewidth]{ 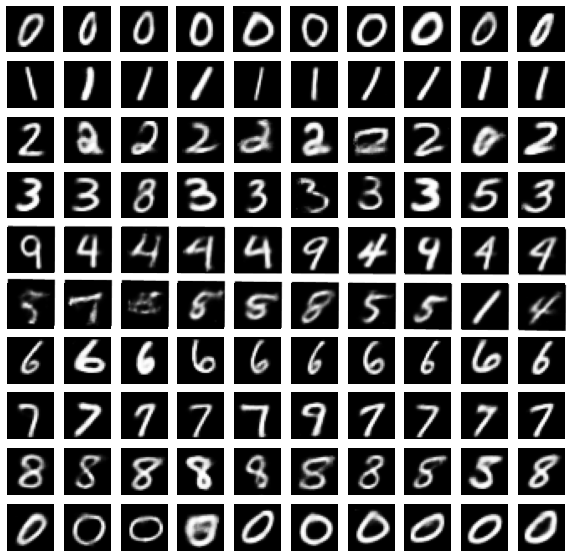}
\end{minipage}
\begin{minipage}{.23\textwidth}
  \centering
  \includegraphics[width=\linewidth]{ 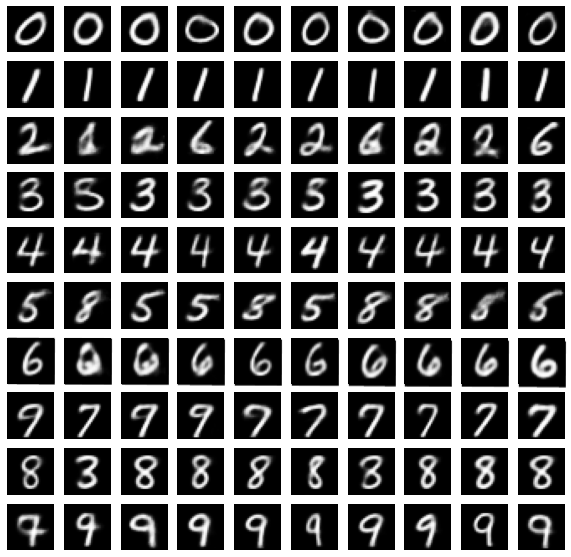}
\end{minipage}
    \caption{Samples from unsupervised training on MNIST. Each component (row) with 10 samples (column) GMVAE (left) vs. Binarized GMVAE (right)}
    \label{fig:test_gm}
\end{figure}

Generative replay requires the fit-sample-replay loop~\cite{shin2017continual,rostami2019complementary,rostami2020generative}, which highly depends on the generated quality of the samples. First, we tested the compatibility of the Gaussian mixture prior to binarized regularization, since it is commonly used for conditional replay and proves to be effective. We test a GMVAE with a binarized GMVAE on MNIST (see Figure \ref{fig:test_gm}). We have compared state-of-the-art CL methods that use generative replay to mitigate catastrophic forgetting. We observe that binarized GMVAE can better disentangle the classes with fewer ambiguities. 

Then we test the binarized generative replay models on the most challenging incremental-class learning scenario. We performed experiments using generative replay(GR) and brain-inspired generative replay (BIR). We conclude that with our modification, the forgetting effect of early tasks can be generally reduced, under single modal (Gaussian) and multimodal (GMM) latent representation, with or without conditional replays (see Table \ref{table:bir}).  

\begin{figure}[h]
    \centering
\begin{minipage}{.23\textwidth}
  \centering
  \includegraphics[width=\linewidth]{ 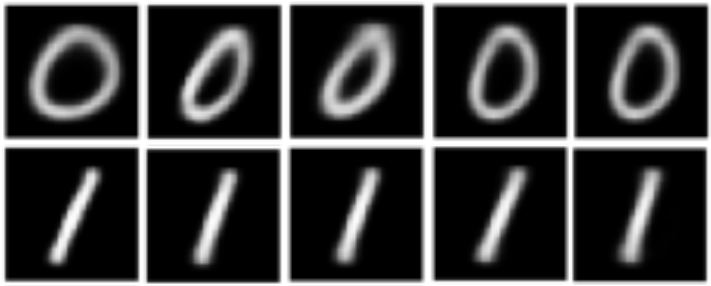}
  \label{fig:test1}
\end{minipage}
\begin{minipage}{.23\textwidth}
  \centering
  \includegraphics[width=\linewidth]{ 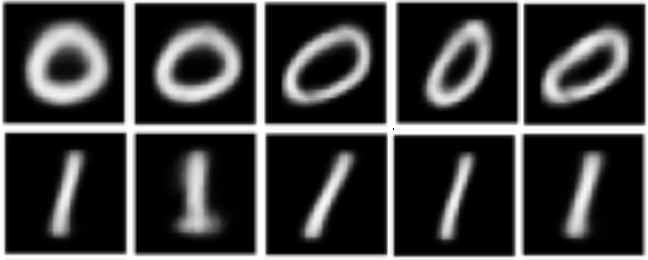}
  \label{fig:test2}
\end{minipage}
    \caption{Conditional Samplings from task 1 and task 2 after task 10. Brain-inspired Replay (GMVAE) vs. Brain-inspired Replay(B-GMVAE, r = 0.01). Binarized prior shows higher variances in replayed early tasks.}
    \label{fig:sample_cl}
\end{figure}

In Figure~\ref{fig:sample_cl}, we observe that the variance in earlier tasks is maintained better, which results in less forgetting of earlier tasks which is an important challenge. The worse performance in more recent tasks and better performance on average suggests that the improvement is not from the ability to reconstruct better samples in general, but from maintaining more diverse knowledge in the model. This observation indicates that we have achieved learning more general and robust factors in the latent space.

\subsection{Ablation Experiments}

We study the effect of major hyperparameters on the quality of reconstrcuted samples.

\subsubsection{Behavior with different $r$ and latent dimension}
\begin{figure}[ht]
    \centering
    \includegraphics[width=0.9\linewidth]{ 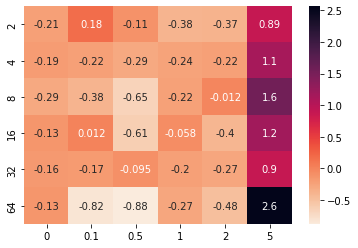}
    \caption{Normalized reconstruction error with different latent dimension (verticle axis) and $r$ (horizontal axis).  Smaller values(lighter color)denote better reconstruction performance.}
    \label{fig:heat_recon}
\end{figure}

\begin{figure}[ht]
    \centering
    \includegraphics[width=0.9\linewidth]{ 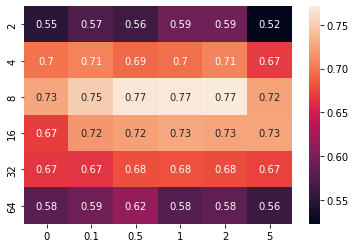}
    \caption{Normalized mutual information with different latent dimension(verticle axis) and $r$(horizontal axis). Larger values(lighter color)denote better clustering performance.}
    \label{fig:heat_nmi}
   
\end{figure}

We   study how different latent sizes $d_z$ and different $r$ can affect the reconstruction and disentanglement quality using the MNIST dataset. Results are presented in Figures \ref{fig:heat_recon} and \ref{fig:heat_nmi}. We observe see that a small $r([0.1,2])$ can improve both the reconstruction and disentanglement quality. A large $r(>2)$ can impact the reconstruction while staying on par for unsupervised clustering with the original model.

 \begin{table*}[htb]
\begin{adjustbox}{width=\textwidth,center}
\begin{tabular}{|l|l|lllllllllll|}
\hline
Dataset                         & Model                & Task 1                                 & 2                                      & 3                                      & 4                                      & 5                                      & 6                                      & 7                                      & 8                                      & 9               & 10                                    & Average                                \\ \hline
                                & GR                   & 0.7531                                 & 0.7771                                 & \textbf{0.8382}                        & 0.8535                                 & 0.4185                                 & 0.8262                                 & \textbf{0.9697}                        & \textbf{0.8930}                        & \textbf{0.9466} & \textbf{0.9832}                       & 0.8259                                 \\
                                & Binarized(r =0.01)        & {\color[HTML]{3166FF} \textbf{0.8041}} & {\color[HTML]{3166FF} \textbf{0.8308}} & {\color[HTML]{FE0000} 0.8198}          & {\color[HTML]{3166FF} \textbf{0.8545}} & {\color[HTML]{3166FF} \textbf{0.6212}} & {\color[HTML]{3166FF} \textbf{0.8823}} & {\color[HTML]{FE0000} 0.9635}          & 0.8298                                 & 0.9384          & 0.9752                                & {\color[HTML]{3166FF} \textbf{0.8519}} \\ \cline{2-13} 
                                & BIR(G)               & \textbf{0.7673}                        & 0.7780                                 & 0.7316                                 & 0.7535                                 & 0.6874                                 & 0.8442                                 & 0.9676                                 & \textbf{0.9377}                        & \textbf{0.9682} & 0.9822                                & 0.8418                                 \\
                                & Binarized(r =0.01)    & {\color[HTML]{FE0000} 0.5112}          & {\color[HTML]{3166FF} \textbf{0.8943}} & {\color[HTML]{3166FF} \textbf{0.7955}} & {\color[HTML]{3166FF} \textbf{0.7802}} & {\color[HTML]{3166FF} \textbf{0.7760}} & {\color[HTML]{3166FF} \textbf{0.9092}} & {\color[HTML]{3166FF} \textbf{0.9729}} & 0.9270                                 & 0.9569          & \textbf{0.9851}                       & {\color[HTML]{3166FF} \textbf{0.8508}} \\ \cline{2-13} 
                                & BIR(GMM)             & 0.8918                                 & 0.8890                                 & \textbf{0.8062}                        & 0.8149                                 & 0.8024                                 & 0.9215                                 & \textbf{0.9833}                        & 0.9660                                 & \textbf{0.9630} & \textbf{0.9911}                       & 0.9029                                 \\
\multirow{-6}{*}{splitMNIST}    & Binarized(r =0.002) & {\color[HTML]{3166FF} \textbf{0.9082}} & {\color[HTML]{3166FF} \textbf{0.9665}} & {\color[HTML]{FE0000} 0.8043}          & {\color[HTML]{3166FF} \textbf{0.8564}} & {\color[HTML]{3166FF} \textbf{0.8371}} & {\color[HTML]{3166FF} \textbf{0.9417}} & {\color[HTML]{FE0000} 0.9749}          & 0.9660                                 & 0.9610          & 0.9881                                & {\color[HTML]{3166FF} \textbf{0.9204}} \\ \hline
                                & GR                   & 0.9447                                 & \textbf{0.9508}                        & 0.9188                                 & 0.9104                                 & 0.8738                                 & \textbf{0.9081}                        & \textbf{0.8993}                        & 0.8981                                 & 0.9187          & 0.9299                                & 0.9153                                 \\
                                & Binarized(R=0.0002)       & {\color[HTML]{3166FF} \textbf{0.9508}} & {\color[HTML]{FE0000} 0.9407}          & {\color[HTML]{3166FF} \textbf{0.9227}} & {\color[HTML]{3166FF} \textbf{0.9165}} & {\color[HTML]{3166FF} \textbf{0.9003}} & {\color[HTML]{FE0000} 0.9017}          & {\color[HTML]{FE0000} 0.8723}          & \textbf{0.9155}                        & \textbf{0.9223} & \textbf{0.9334}                       & {\color[HTML]{3166FF} \textbf{0.9176}} \\ \cline{2-13} 
                                & BIR(G)               & 0.9495                                 & 0.9689                                 & 0.9724                                 & 0.9734                                 & \textbf{0.9757}                        & 0.9736                                 & 0.9748                                 & {\color[HTML]{333333} \textbf{0.9749}} & \textbf{0.9739} & \textbf{0.9751}                       & 0.9712                                 \\
                                & Binarized(r=0.0001)   & {\color[HTML]{3166FF} \textbf{0.9512}} & {\color[HTML]{3166FF} \textbf{0.9692}} & {\color[HTML]{3166FF} \textbf{0.9736}} & {\color[HTML]{3166FF} \textbf{0.9745}} & {\color[HTML]{FE0000} 0.9756}          & {\color[HTML]{3166FF} \textbf{0.9743}} & {\color[HTML]{3166FF} \textbf{0.9754}} & 0.9735                                 & 0.9730          & 0.9748                                & {\color[HTML]{3166FF} \textbf{0.9715}} \\ \cline{2-13} 
                                & BIR(GMM)             & 0.9650                                 & 0.9679                                 & \textbf{0.9708}                        & 0.9737                                 & 0.9750                                 & 0.9756                                 & 0.9764                                 & \textbf{0.9766}                        & \textbf{0.9783} & 0.9789                                & 0.9739                                 \\
\multirow{-6}{*}{permutedMNIST} & Binarized(r= 0.01)     & {\color[HTML]{3166FF} \textbf{0.9652}} & {\color[HTML]{3166FF} \textbf{0.9699}} & {\color[HTML]{FE0000} 0.9695}          & {\color[HTML]{3166FF} \textbf{0.9750}} & {\color[HTML]{3166FF} \textbf{0.9751}} & {\color[HTML]{3166FF} \textbf{0.9769}} & {\color[HTML]{3166FF} \textbf{0.9781}} & 0.9765                                 & 0.9776          & \textbf{0.9791}                       & {\color[HTML]{3166FF} \textbf{0.9743}} \\ \hline
                                & GR                   & 0                                      & 0                                      & 0                                      & 0                                      & 0                                      & 0.003                                  & 0.016                                  & 0.014                                  & 0.1520          & \textbf{0.4370}                       & 0.0622                                 \\
                                & GR(r=0.001)          & 0                                      & 0                                      & {\color[HTML]{3166FF} \textbf{0.003}}  & {\color[HTML]{3166FF} \textbf{0.001}}  & 0                                      & {\color[HTML]{3166FF} \textbf{0.006}}  & {\color[HTML]{3166FF} \textbf{0.023}}  & \textbf{0.032}                         & \textbf{0.185}  & 0.413                                 & {\color[HTML]{3166FF} \textbf{0.0663}} \\ \cline{2-13} 
                                & BIR                  & 0.016                                  & 0.039                                  & 0.051                                  & 0.058                                  & 0.064                                  & 0.095                                  & 0.199                                  & \textbf{0.212}                         & \textbf{0.403}  & {\color[HTML]{000000} \textbf{0.673}} & 0.1810                                 \\
                                & Binarized(r = 0.01)      & {\color[HTML]{3166FF} \textbf{0.022}}  & {\color[HTML]{3166FF} \textbf{0.047}}  & 0.051                                  & {\color[HTML]{3166FF} \textbf{0.075}}  & {\color[HTML]{3166FF} \textbf{0.084}}  & {\color[HTML]{3166FF} \textbf{0.102}}  & {\color[HTML]{3166FF} \textbf{0.254}}  & 0.204                                  & 0.395           & 0.668                                 & {\color[HTML]{3166FF} \textbf{0.1902}} \\ \cline{2-13} 
                                & BIR(GMM)             & 0.034                                  & 0.037                                  & 0.060                                  & 0.088                                  & \textbf{0.108}                         & 0.129                                  & 0.263                                  & \textbf{0.241}                         & \textbf{0.458}  & \textbf{0.672}                        & 0.2090                                 \\
\multirow{-6}{*}{CIFAR100}      & Binarized(r =0.01)   & {\color[HTML]{3166FF} \textbf{0.039}}  & {\color[HTML]{3166FF} \textbf{0.052}}  & {\color[HTML]{3166FF} \textbf{0.061}}  & {\color[HTML]{3166FF} \textbf{0.104}}  & {\color[HTML]{FE0000} 0.096}           & {\color[HTML]{3166FF} \textbf{0.153}}  & {\color[HTML]{3166FF} \textbf{0.300}}  & 0.238                                  & 0.453           & 0.665                                 & {\color[HTML]{3166FF} \textbf{0.2161}} \\ \hline
\end{tabular}
\end{adjustbox}
\label{table:bir}
\caption{Incremental-class learning accuracy from task 1 to 10. Generative Replay vs. Brain-inspired Generative Replay(BIR) vs. Binarized versions. BIR(G/GMM) specifies the prior for conditional replay is a unimodal Gaussian(G) or a Gaussian mixture(GMM). Better performance in bold. Better performance in earlier tasks(1-7) and in average(Average)is in: blue if it's a binarized model; red if it's an original model.}

\end{table*}

\textbf{Choosing r}:
The hyperparameter $r$ decides how disentangled the binarized priors are in the latent embedding space. The center (mean) distance between the binarized priors in a $d$-dimensional space has range the $[2r, 2r\sqrt{d}]$. We can see at large $d$, with a low learning rate, a large $r$ will make it difficult for a latent distribution to shift to a distant mean when new representations are learned. This is undesirable at the beginning of training or for learning high-level concepts subject to large changes. Hence, a small $r$ works well in practice. Typically a smaller $r$ is required for a more complex dataset, lower learning rate, and higher latent dimension. For our experiments, ($r \in [0.5,2]$ for MNIST, $r \in [0.1,1]$ for CelebA, and $r \in [0.0001,0.01]$ for continual learning). (CL representations are subject to drastic changes, so $r$ needs to be extremely small)
\subsubsection{Source of improvement}

We empirically investigate why using our regularization techniques leads to improved performance.  

\textbf{Not from Exponential growth of high-density area:} Since our method also employs the idea to exponentially increase the high-density size like hyperspherical-VAE (NVAE) as a way to resolve the issues with Gaussian distribution. We compare a binarized VAE and NVAE in table \ref{table:MNIST}, and observe a significant performance improvement compared to NVAE.
\textbf{No extra resource required:} In table \ref{table:MNIST}, we also compared the time cost to train a binarized model. We observed that binarized VAE does not increase the time, and in reality, there is a $\approx1\%$ performance increase for binarized toyVAEs. This might be due to more minor variance in the binarized $\mu_{q_{\phi}(z|x)} - \mu_{p(z)}$ matrix, resulting in better floating-number matrix multiplication in the program.

\section{Conclusions}
We proposed binarized regularization to improve VAEs. Our approach helps to distribute the distribution mass in the latent embedding space to represent the input distribution in a more distributed way.  Extensive experiments demonstrate that our approach exhibits good properties such as better disentanglement, better reconstruction quality, and better sampling variance. The effectiveness of our model is   exhibited in all cases we tested, with simple modification and no extra cost. The encoding using symmetric bimodal distributions has great potential in applications of VAE, as demonstrated in the case of generative replay for CL. Future work includes applying our approach  to large hierarchical VAE models and other latent space models.
 

\onecolumn

\section*{\centering{\LARGE{Appendix}}}

 Experiments implementations are available at \url{https://github.com/zizhao-hu/BVAE}

\section*{1.Latent Visualizations}
\subsection*{1.1 Pair-wise 4-D Latent Visualizations in 2-D}
Visualization for the latent space(dimension = 4) of VAEs and the binarized versions. Symmetry is shown in the binarized VAEs and posterior collapse does not appear in binarized disentangling $\beta$-VAE. We use MNIST training dataset.
\begin{figure*}[h]
\includegraphics[width=\linewidth]{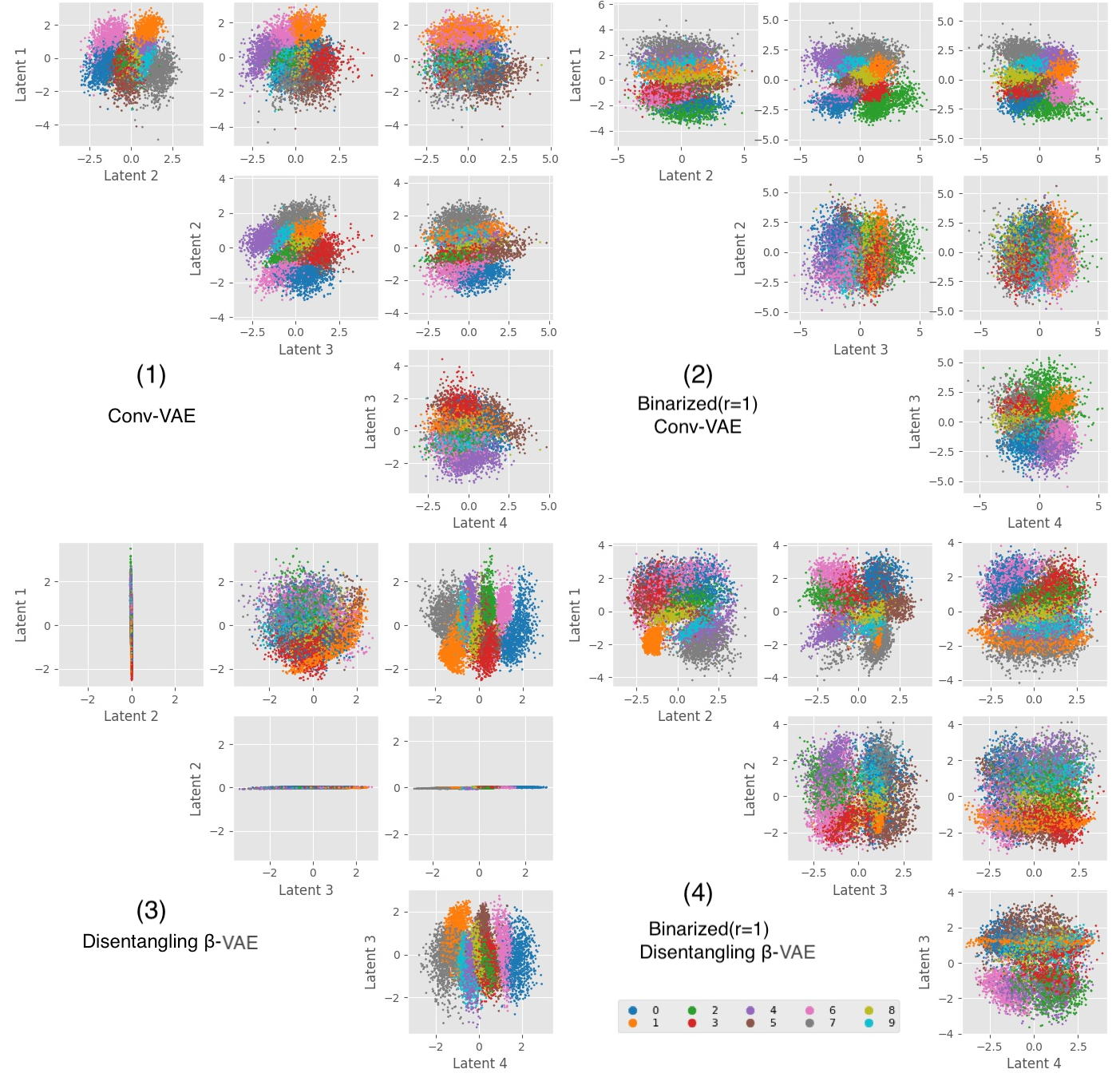}
    \caption{Pair-wise latent space visualization }
    \label{fig:4_latent_vis}
\end{figure*}
In row 1, the binarized latent space is more symmetric. The correlation between two latent dimensions is also reduced. The data points are spread out in a larger area without requiring additional training resources.

In row 2, the binarized latent space does not suffer from posterior collapse. 
\section*{2. UMAP 16-D Latent Visualizations in 2-D}
Four models(VAE, $\beta$-VAE,  disentangling  $\beta$-VAE,  GMVAE) with four binarized versions. Column one denotes the original models corresponding to binarizing hyperparameter $r=0$. Column $2,3,4$ correspond to $r=0.1$,$r=1$ and $r=2$. We use MNIST training dataset. The best disentanglement of classes happens at $r = 1$ for all unimodal Gaussian prior models. $r = 0.1$ for Gaussian mixture prior.
\begin{figure*}[h]
    \includegraphics[width=\linewidth]{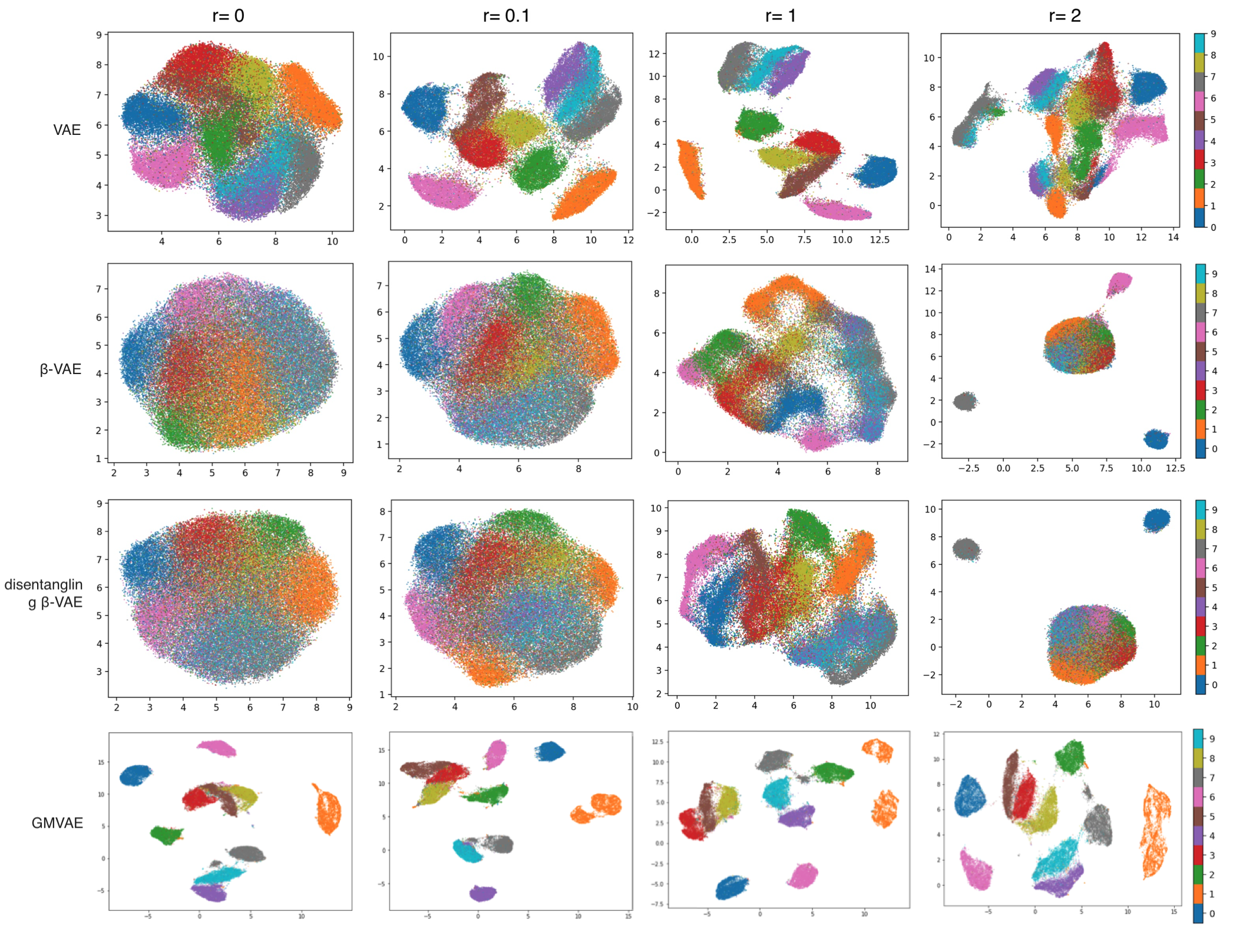}
    \caption{UMAP visualization of the latent manifolds in 2-D.}
    \label{fig:my_label1}
\end{figure*}
In rows 1, 2, and 3, the best disentanglement happens at $r=1$. When $r=2$, the latent is over-disentangled.

In row 4, the binarized model detaches the border between classes, and achieves better disentanglement. 

\section*{3. Visualization of Different Hyperparameters}
\subsection*{3.1 Reconstruction}
We test $r=0,0.01,0.1,1$ settings on the CelebA dataset. Reconstruction loss is reduced for binarized models $r=0.01,0.1$. The running time is also slightly reduced(Table \ref{tab:recon_time}).

\begin{table}[ht]
\centering
\begin{tabular}{|l|ll|}
\hline
              & Reconstruction loss & Running time(Average of 50 epochs) \\ \hline
r=0(original) & 0.0137              & 115.26                             \\
r=0.01        & \textbf{0.0128}     & 115.36                             \\
r=0.1         & \textbf{0.0128}     & \textbf{113.32}                    \\
r=1           & 0.0160              & 116.8                              \\ \hline
\end{tabular}
\caption{Reconstruction quality and running time}
\label{tab:recon_time}
\end{table}

\begin{figure}[h]
\centering
\begin{minipage}{.5\textwidth}
    \includegraphics[width=0.99\linewidth]{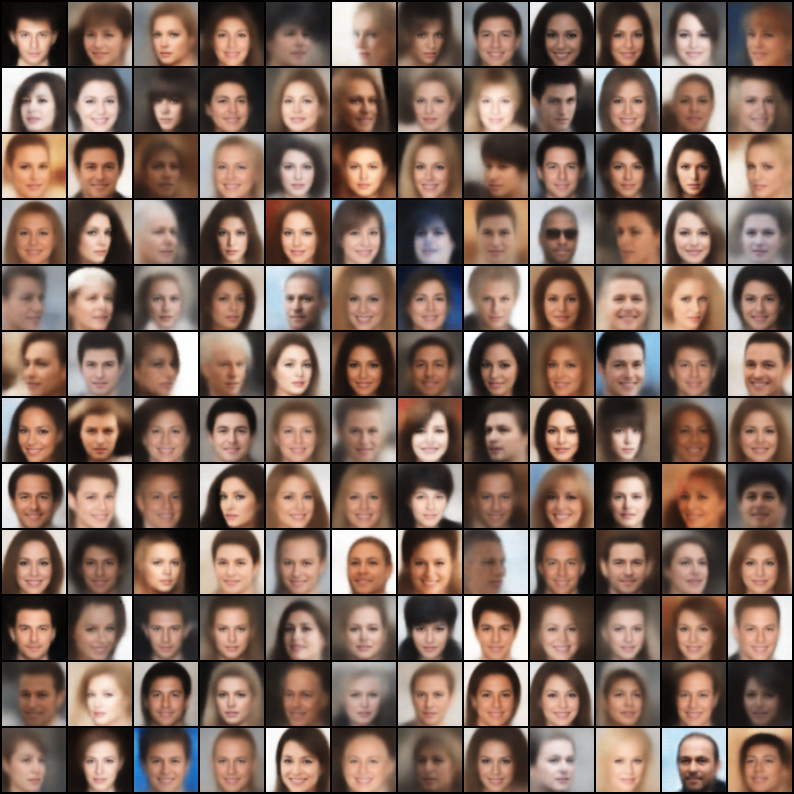}
    \centering
    \\r=0(original)
    \label{fig:my_label2}
\end{minipage}%
\begin{minipage}{.5\textwidth}
    \includegraphics[width=0.99\linewidth]{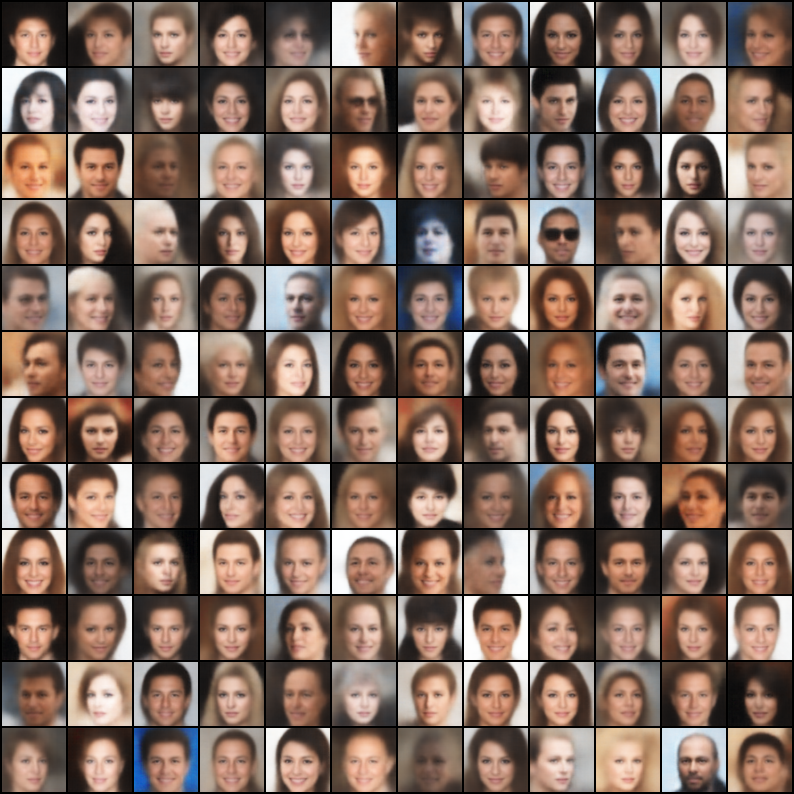}
    \centering
    \\{r=0.01}
    \label{fig:my_label3}
\end{minipage}
\begin{minipage}{.5\textwidth}
    \includegraphics[width=0.99\linewidth]{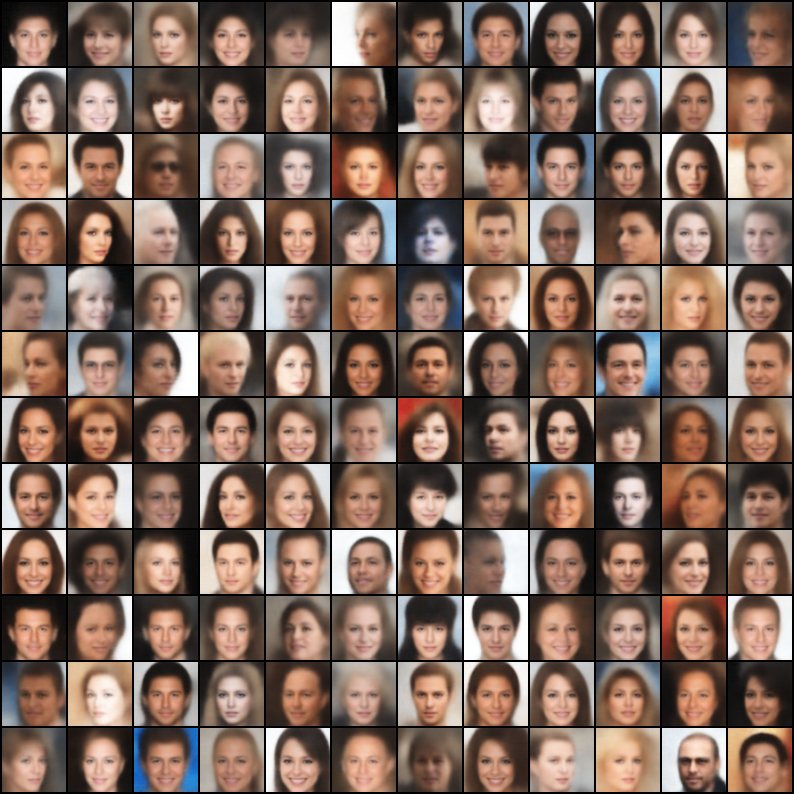}
    \centering
    \\{r=0.1}
    \label{fig:my_label4}
\end{minipage}%
\begin{minipage}{.5\textwidth}
    \includegraphics[width=0.99\linewidth]{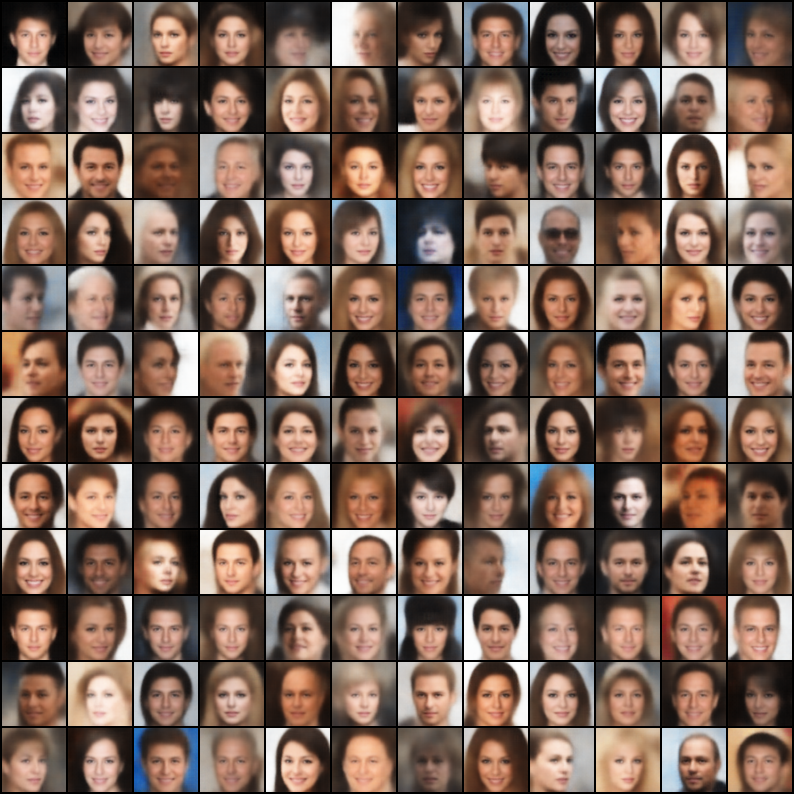}
    \centering
    \\{r=1}
    \label{fig:my_label5}
\end{minipage}
 \caption{Reconstruction with different r settings}
\end{figure}
\subsection*{3.2 Sampling}
Sampling is conducted on the same models as in the last section. We use the same latent space vector to generate the image in all four models. We can see that when $r$ is small, the samples have more distinct facial features, sharper edges, and  vibrant colors. When $r$ is large, the samples sampled from the original prior are from the ``ambiguous region", since the sampling prior is between the binarized centers. And the generated images are ambiguous and lack distinction as expected. 
\begin{figure}[h]
\centering
\begin{minipage}{.5\textwidth}
    \includegraphics[width=0.99\linewidth]{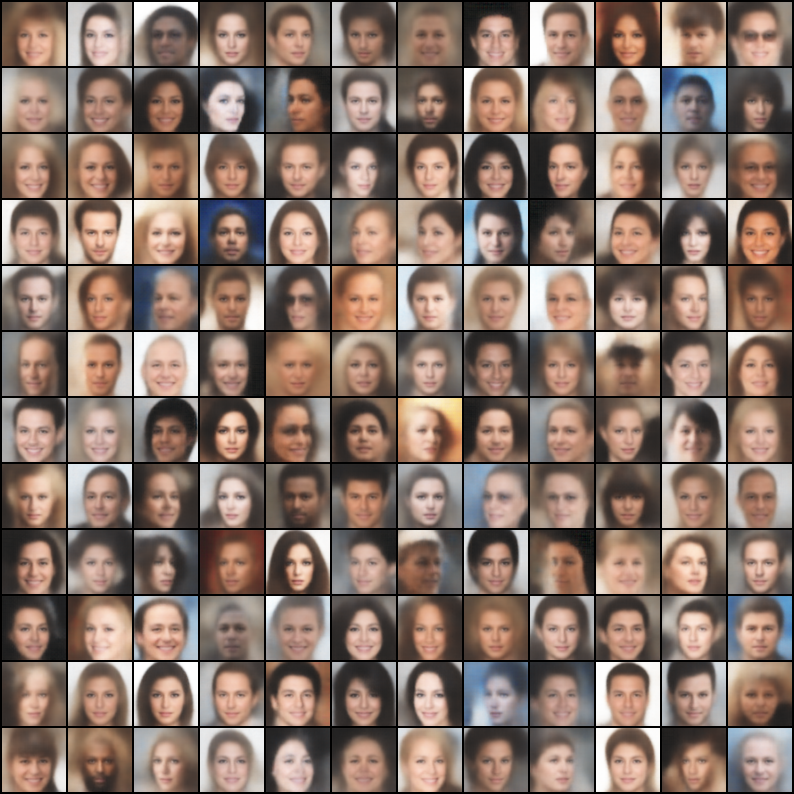}
    \centering
    \\r=0(original)
    \label{fig:my_label6}
\end{minipage}%
\begin{minipage}{.5\textwidth}
    \includegraphics[width=0.99\linewidth]{ 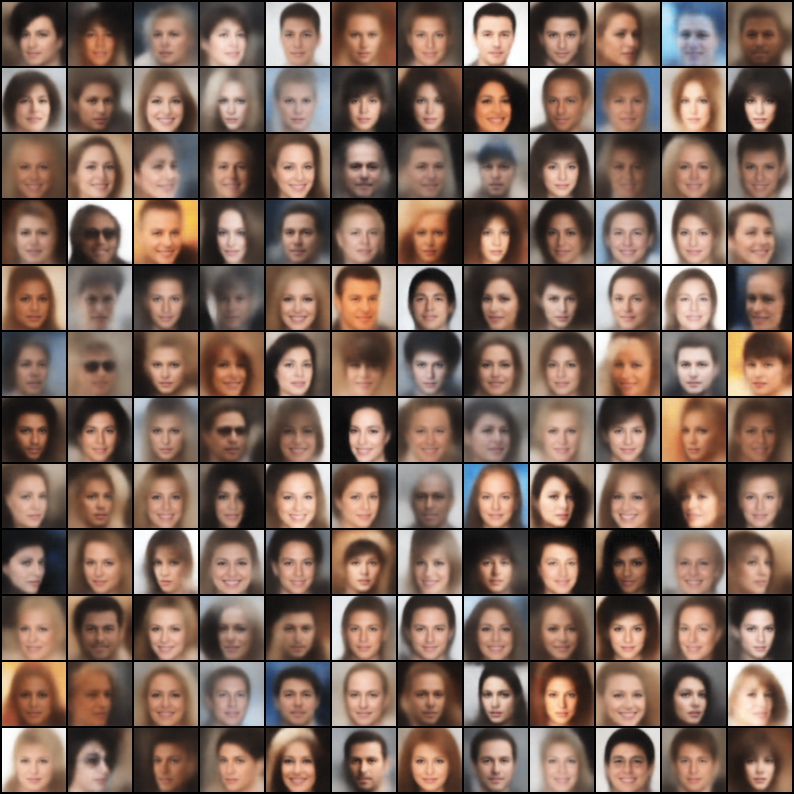}
    \centering
    \\{r=0.01}
    \label{fig:my_label7}
\end{minipage}
\begin{minipage}{.5\textwidth}
    \includegraphics[width=0.99\linewidth]{ 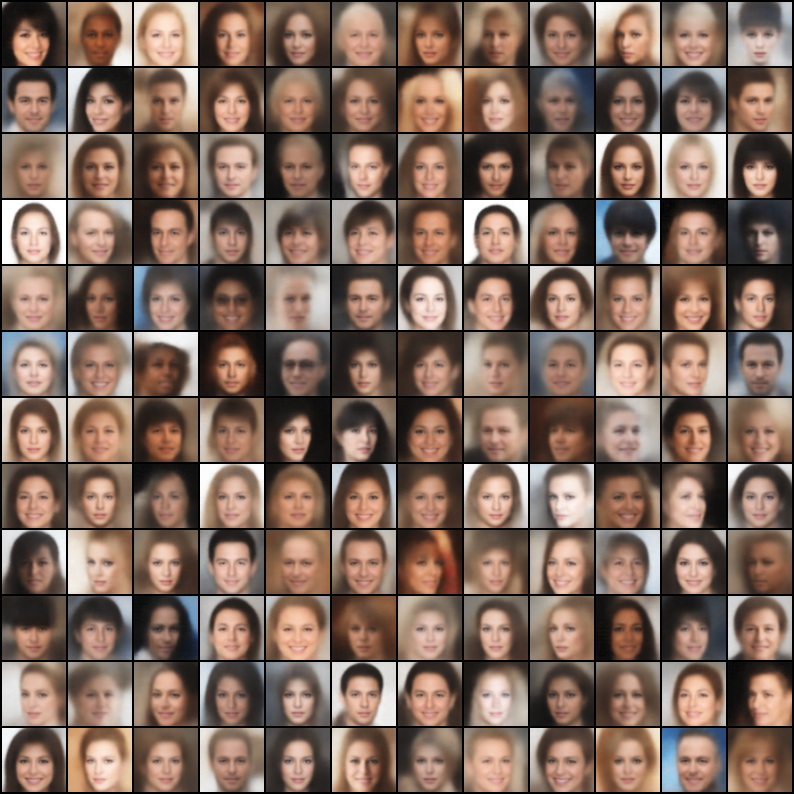}
    \centering
    \\{r=0.1}
    \label{fig:my_label8}
\end{minipage}%
\begin{minipage}{.5\textwidth}
    \includegraphics[width=0.99\linewidth]{ 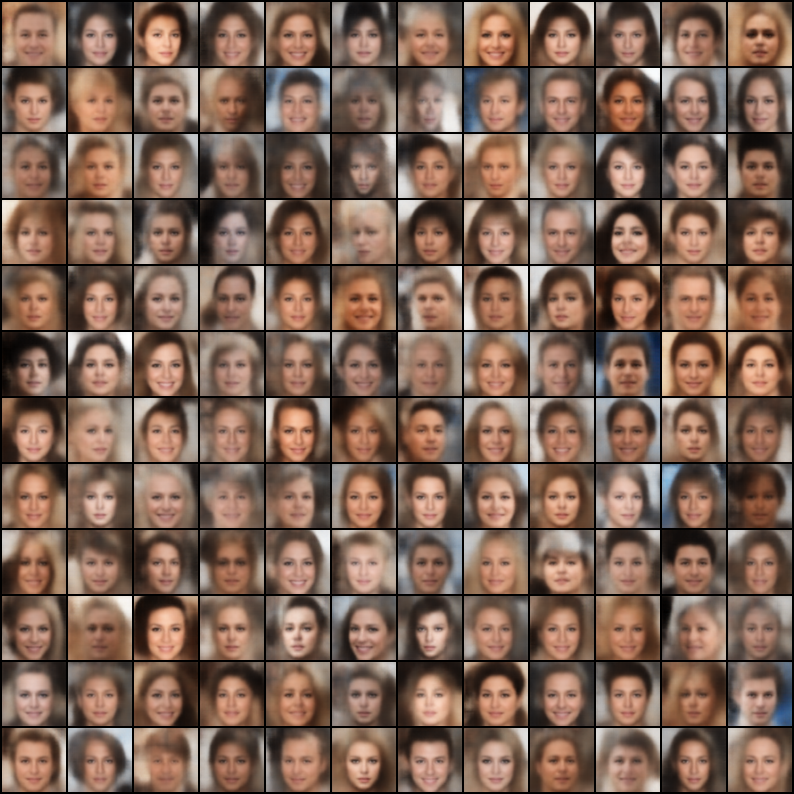}
    \centering
    \\{r=1}
    \label{fig:my_label9}
\end{minipage}
 \caption{Reconstruction with different r settings}
\end{figure}
\section*{4. Reduced-bias Sampling for High-variance Posteriors}
\subsection*{4.1 Binarized approximate posterior}
\begin{figure}[h]
    \centering
    \includegraphics[width = 0.76\linewidth]{ 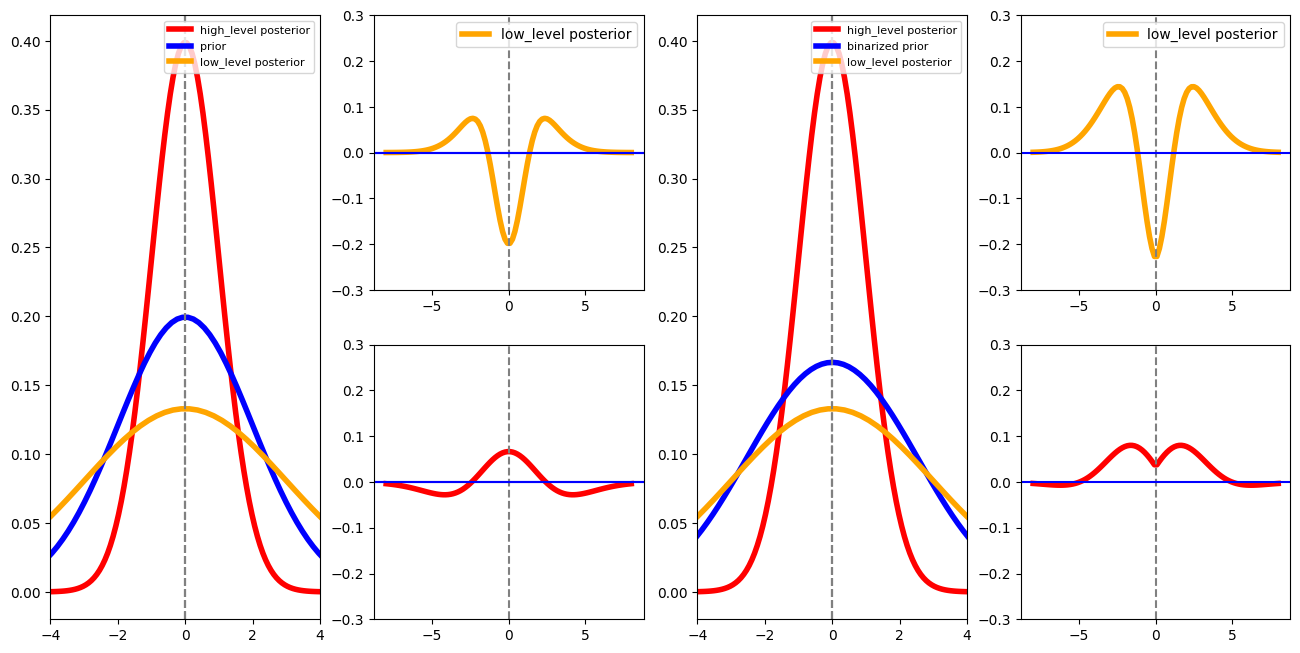}
    \caption{Estimation and sampling distributions}
    \label{fig:re_bias}
\end{figure}

Optimal approximate posterior estimation for VAE with a Gaussian prior $ p_{\phi}(z|x)\sim \mathcal{N}(\mu,\sigma^2)$, assuming some error $\epsilon_\mu$ and $\epsilon_\sigma$
\begin{equation}
    p_{\phi}(z|x)\sim \mathcal{N}(\mu+\epsilon_\mu,(\sigma+\epsilon_\sigma)^2) = \frac{1}{(\sigma+\epsilon_\sigma)\sqrt{2\pi}} 
  \exp\left( -\frac{1}{2}\left(\frac{x-\mu-\epsilon_\mu}{(\sigma+\epsilon_\sigma)}\right)^{\!2}\,\right)
\end{equation}
For binarized VAE: 
\begin{equation}
p_{\phi}(z|x)_{Binarized}\sim \mathcal{N}(|\mu+\epsilon_\sigma|-r,(\sigma+\epsilon_\sigma)^2) = \frac{1}{(\sigma+\epsilon_\sigma)\sqrt{2\pi}} 
  \exp\left( -\frac{1}{2}\frac{(|x-\mu-\epsilon_\mu|-r)^2}{(\sigma+\epsilon_\sigma)^2}\,\right)
  \end{equation}
Difference:
\begin{equation}
 p_{\phi}(z|x) - p_{\phi}(z|x)_{Binarized} = \frac{1}{(\sigma+\epsilon_\sigma)\sqrt{2\pi}} 
  \left(\exp\left( -\frac{1}{2}\frac{(x-\mu-\epsilon_\mu)^2}{(\sigma+\epsilon_\sigma)^2}\,\right)-\exp\left( -\frac{1}{2}\frac{(x-\mu-\epsilon_\mu)^2-2r|x-\mu-\epsilon_\mu|+r^2}{(\sigma+\epsilon_\sigma)^2}\,\right) \right)
\end{equation}
When $x\approx\mu$ we have $(x-\mu-\epsilon_\mu)^2 \approx {\epsilon_\mu}^2<<(\sigma+\epsilon_\sigma)^2$, when $\sigma$ is sufficiently large, then approximate by first-order Taylor expansion:

\begin{equation}
\begin{aligned}
 p_{\phi}(z|x) - p_{\phi}(z|x)_{Binarized}&\approx \frac{1}{(\sigma+\epsilon_\sigma)\sqrt{2\pi}} \left(1 -\frac{1}{2}\frac{(x-\mu-\epsilon_\mu)^2}{(\sigma+\epsilon_\sigma)^2} -1 + \frac{1}{2}\frac{(x-\mu-\epsilon_\mu)^2-2r|x-\mu-\epsilon_\mu|+r^2}{(\sigma+\epsilon_\sigma)^2}\,\right)\\[1em]
&= \frac{-2r|x-\mu-\epsilon_\mu|+r^2}{2(\sigma+\epsilon_\sigma)^3\sqrt{2\pi}}
 \end{aligned}
\end{equation}
Thus, we have
\begin{equation}
  p_{\phi}(z|x) > p_{\phi}(z|x)_{Binarized} \; \text{when} \; |x-\mu-\epsilon_\mu|<\frac{r}{2}\; , \; x\approx\mu \; ,\text{and} \; \sigma \; \text{is sufficiently large}
\end{equation}
This allows the binarized approximate posterior to be less distributed around the mean. When sampled using the same prior distribution, reduced bias is thus achieved. (Also see Figure \ref{fig:re_bias}. The red curve denotes high-variance posterior. Column 1 is a scenario when the estimated posterior has a larger or smaller variance than the prior. Column 2 is the difference between the posteriors and the prior. Columns 3, and 4 are binarized models.)

\end{document}